\documentclass{article}
\usepackage{spconf,amsmath,graphicx,hyperref}
\usepackage{booktabs}
\usepackage{cite}
\usepackage{multirow}
\usepackage{xcolor}
\usepackage{amssymb}
\usepackage{pifont}
\makeatletter
\let\originalthebibliography\thebibliography
\renewcommand{\thebibliography}[1]{%
  \originalthebibliography{#1}%
  \fontsize{9}{10}\selectfont
}
\makeatother

\title{X2-Turn: Frame-Synchronous Dual-Head Modeling for Joint Streaming ASR and Turn State Prediction}
\name{Kaiqi Fu, Rime Wen, Altman Lin, Shawn Qin, Roy Gan, Hao Wang, Qian Wang}
\address{X Square Robot\\
\texttt{\{kaiqifu, wanghao\}@x2robot.com}}

\begin{document}

\maketitle

\begin{abstract}
Accurate and responsive turn-taking is essential for spoken dialogue systems, which must distinguish in real time between user interruptions, backchannels that should be ignored, and the completion of an utterance. Prior modular approaches typically optimize turn state prediction at the utterance or fixed-chunk level, creating a mismatch with the continuous turn state estimate, and often depend on an auxiliary ASR model, which limits responsiveness and increases overall system complexity. Therefore, we present X2-Turn~\footnote{\scriptsize{Code and models: \url{https://github.com/X-Square-Robot/X2-Turn}.}}, a frame-synchronous turn state prediction method via delayed-stream modeling. Specifically, building on the pretrained Voxtral Realtime model, we introduce a frame-synchronous turn state head that operates in parallel with the ASR head on shared streaming representations, jointly predicting ASR tokens and fine-grained turn states at the frame level. Experiments on bilingual EasyTurn and Full-Duplex-Bench demonstrate that the proposed method achieves an effective trade-off between turn state accuracy and decision latency.
\end{abstract}


\section{Introduction}
Achieving natural spoken dialogue requires systems to seamlessly handle continuous speech, backchannels, and user interruptions while maintaining low latency~\cite{review}. To manage these complex conversational dynamics, a responsive system must continuously estimate fine-grained turn states. These states serve as the foundation for real-time dialogue control, determining when to interrupt text-to-speech (TTS) playback, take the conversational floor, or ignore a user backchannel.

Existing approaches broadly follow either end-to-end or cascaded paradigms. End-to-end full-duplex models~\cite{dgslm,moshi,freeze_omni,omniflatten,personaplex} jointly learn speech understanding, interaction timing, and response generation. For example, Moshi employs a dual-stream architecture that jointly generates text and audio tokens for both speakers, using an inner-monologue mechanism to align semantic and acoustic generation~\cite{moshi}. Although these systems directly model synchronous interactions, jointly optimizing all components on limited full-duplex data may constrain the scale and general capabilities of the dialogue backbone.

Cascaded approaches instead decouple interaction control from response generation~\cite{flexduo,easyturn,jalturn,fastturn,soulx,joyaitalker}. A representative example is the VAD--ASR--turn-detection pipeline~\cite{tenvad}, in which a front-end voice activity detection (VAD) module first segments the continuous user speech stream, an ASR model transcribes each resulting segment, and a semantic turn-detection model determines the turn state from the transcript. Because each stage depends on the output of the preceding stage, this pipeline introduces sequential latency and error propagation. To reduce the dependence on a separate upstream ASR module, subsequent studies have explored tighter integration between ASR and turn detection~\cite{easyturn,fastturn,jalturn,soulx}. EasyTurn jointly predicts transcriptions and four turn states from VAD-segmented utterances, replacing the separate downstream turn-detection model with a unified model that reasons over ASR-derived transcripts~\cite{easyturn}. JAL-Turn combines frozen SenseVoice and CPC representations to classify hold and shift states at candidate boundaries, whereas FastTurn integrates partial CTC hypotheses with acoustic cues to make low-latency decisions as the transcript is incrementally updated, without waiting for utterance completion~\cite{jalturn,fastturn}. SoulX-Duplug further interleaves chunk-level ASR and state tokens within a single autoregressive stream, achieving strong turn-taking performance. However, it still relies on an external ASR model to guide state prediction during inference~\cite{soulx}. 
Despite these advances, these methods generally operate at the utterance or chunk level rather than continuously estimating the turn state at every frame. This mismatch in temporal granularity limits their responsiveness in real-time interactions.

More recently, Voxtral Realtime introduced a natively streaming ASR architecture based on delayed-stream modeling~\cite{dsm,moshi}. It emits transcription tokens synchronously with the input audio at a fixed frame rate of 80\,ms~\cite{voxtral}. Inspired by this architecture, we introduce a turn state prediction head parallel to the ASR head, with both heads jointly optimized over shared causal decoder representations to predict ASR tokens and turn states simultaneously. To temporally align the two tasks, we propose ASR-anchored supervision, which projects word-level turn annotations onto the frame-level positions of the corresponding ASR tokens. Experiments on the Chinese and English EasyTurn test sets demonstrate that the proposed method achieves an effective trade-off between turn state accuracy and decision latency.

\begin{figure*}[t]
\centering

\includegraphics[width=0.9\textwidth]{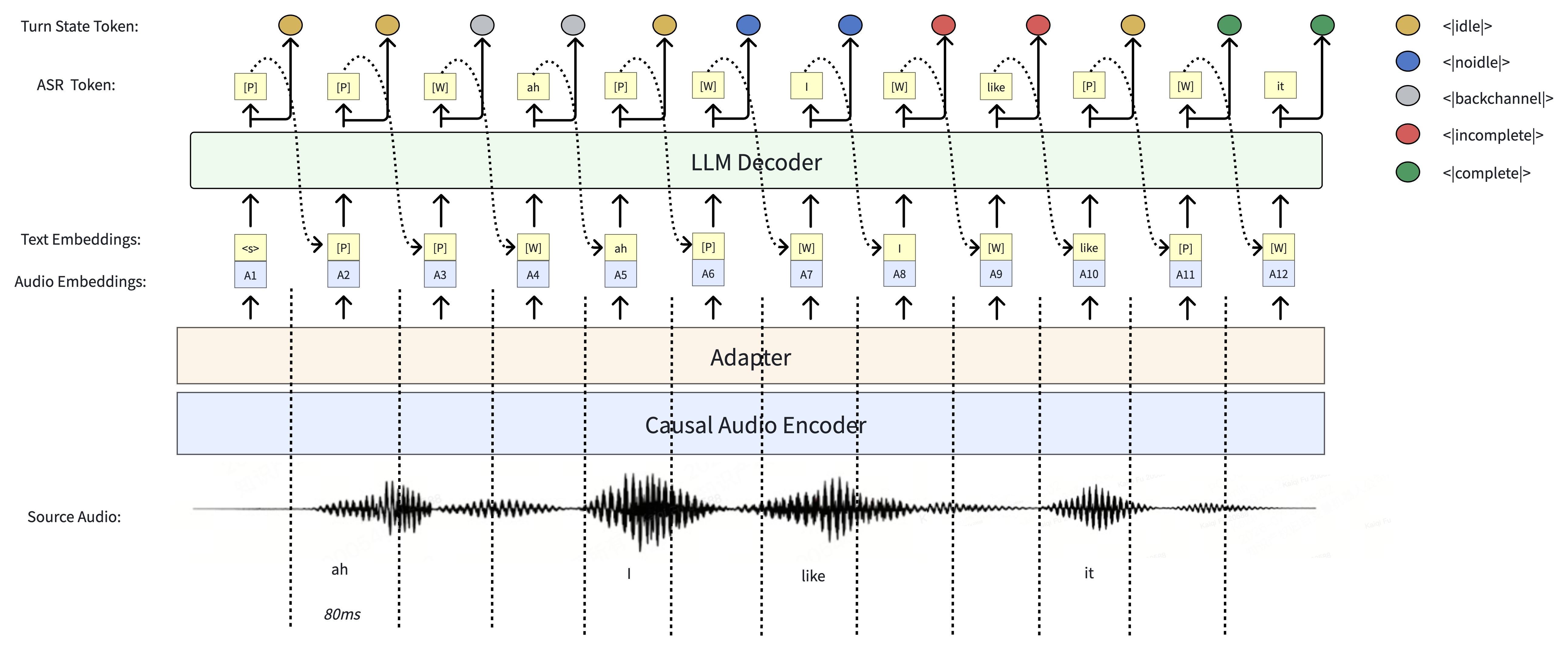}
\caption{Overview of X2-Turn, the proposed frame-synchronous dual-head architecture, with a target delay of $\tau=80\,\mathrm{ms}$. Once the onset of a word has been observed and the target delay has elapsed, the ASR head emits a word-boundary token \texttt{[W]}, while the turn state head simultaneously predicts the corresponding state. Subsequent subword tokens are then emitted frame by frame.}
\label{fig:method}
\end{figure*}

Our main contributions are summarized as follows:
\begin{enumerate}
    \item We propose X2-Turn, which extends a pretrained delayed-stream ASR model with a parallel turn state head, enabling joint frame-synchronous ASR and turn state prediction within a single streaming forward pass.
    \item We design a unified turn state label set that supports interruption, turn completion, and backchannel detection. We further introduce an ASR-anchored supervision method that projects word-level turn annotations onto the frame-level ASR token timeline.
    \item We conduct bilingual experiments on EasyTurn and Full-Duplex-Bench~\cite{fdb,fdbv1.5}, validating the effectiveness of the proposed method for streaming turn state prediction under different latency settings controlled by $\tau$.

\end{enumerate}

\section{Method}
In this section, we first introduce the streaming ASR backbone and dual-head architecture, and then describe the turn state token design and the construction of ASR-anchored turn state labels. An overview of the framework is shown in Fig.~\ref{fig:method}.

\subsection{Dual-Head Modeling}
Our goal is to estimate turn states synchronously with ASR transcription as speech unfolds, while keeping interaction control independent of the downstream dialogue model. To this end, we build our system on Voxtral Realtime~\cite{voxtral}, a natively streaming ASR model based on delayed-stream modeling (DSM). Voxtral Realtime maps an input waveform to a delayed token stream using a causal audio encoder and a language decoder. We retain the backbone architecture and introduce a parallel turn state prediction head. This design decouples turn state estimation from response generation, allowing the resulting turn-taking module to be integrated into different dialogue systems.

The architecture of the proposed method is illustrated in Fig.~\ref{fig:method}. We first briefly review the Voxtral Realtime backbone before presenting our dual-head extension.
Voxtral Realtime consists of three main components: a causal audio encoder that maps 16-kHz waveforms to frame-level audio features, a temporal adapter that downsamples the encoder features to 12.5\,Hz, and a decoder-only language model that emits one token at each 80-ms step. At step $i$, the decoder takes as input the sum of the current audio embedding and the text embedding of the previously emitted ASR token $y_{i-1}^{\mathrm{asr}}$. It then produces a hidden state $h_i$, from which the ASR head predicts the next token. The ASR head is trained using token-level cross-entropy over the ASR vocabulary:

\begin{equation}
\mathcal{L}_{\mathrm{asr}} = -\sum_{i=1}^{T} \log P_{\mathrm{asr}}\left(y_i^{\mathrm{asr}} \mid x_{\le i}, y_{<i}^{\mathrm{asr}}\right),
\end{equation}

where $x_{\leq i}$ denotes the audio observed up to step $i$, $y_{<i}^{\mathrm{asr}}$ denotes the previously emitted ASR tokens, and $T$ is the sequence length.

In addition to ordinary subword tokens, the ASR vocabulary contains two special symbols: a padding token \texttt{[P]} and a word-boundary token \texttt{[W]}. The output token stream is delayed relative to the input audio by a configurable target delay $\tau$. This delay is conditioned into the decoder through AdaRMSNorm, allowing a single model to operate at delays corresponding to different multiples of 80\,ms.

We preserve the original ASR prediction head and add a parallel turn state prediction head. Both heads operate on the shared hidden state $h_i$, enabling ASR and turn state prediction within a single forward pass. The two heads are jointly optimized using the following objective:

\begin{align}
\mathcal{L}_{\mathrm{turn}} &= -\sum_{i=1}^{T} \log P_{\mathrm{turn}}\left(y_i^{\mathrm{turn}} \mid x_{\le i}, y_{<i}^{\mathrm{asr}}\right), \\
\mathcal{L} &= \mathcal{L}_{\mathrm{asr}} + \lambda \mathcal{L}_{\mathrm{turn}},
\end{align}

where $y_i^{\mathrm{turn}}$ denotes the ground-truth turn state at step $i$, and $\lambda$ controls the contribution of the turn state loss. For notational simplicity, the target delay $\tau$ is omitted from the conditional distributions above ($\tau = 0$).

At inference time, autoregressive decoding is driven solely by the ASR head. At each step, the turn head independently predicts the turn state from the shared hidden state $h_i$, and its prediction is not fed back into the decoding loop. Therefore, turn state prediction errors do not affect subsequent ASR decoding. Both transcription and turn states are produced within the same forward pass, without introducing an additional sequential inference stage and latency.

\subsection{Turn-Taking State Token Design}

We define five turn state tokens to represent the evolving state of a user turn:

\begin{itemize}
    \item \texttt{<|idle|>} represents user silence, corresponding to non-speech segments derived from forced alignment.
    \item \texttt{<|noidle|>} indicates active speech without semantic content (e.g., the initial syllables of an utterance).
    \item \texttt{<|incomplete|>} denotes active speech containing partial semantic content.
    \item \texttt{<|complete|>} signifies active speech with complete semantic content.
    \item \texttt{<|backchannel|>} captures user backchannel signals or filler words (e.g., ``um'', ``ah'').
\end{itemize}

Unlike the definition adopted in prior work~\cite{soulx}, our \texttt{<|noidle|>} state refers specifically to early active speech for which sufficient semantic content has not yet been observed. It does not represent non-speech or background noise. This distinction allows the model to separate true silence from the early portion of an ongoing user utterance.

\subsection{ASR-Anchored Turn State Supervision}

We next describe the construction of paired ASR and turn state targets. We first construct the delayed-stream ASR target sequence. Let the $i$-th word span the time interval $[s_i,e_i]$. Each word is tokenized into one or more subword tokens, for example, multiple BPE tokens for an English word. Regardless of the number of subword tokens, each word is represented by a single word-boundary token \texttt{[W]} followed by its corresponding subword tokens.

Given the onset time $s_i$, the target position $p_i$ of the word-boundary token is computed as
\begin{equation}
p_i = \operatorname{round}\left(\frac{s_i}{\Delta}\right)+n_{\tau},
\end{equation}
where $\Delta=80\,\mathrm{ms}$ is the frame duration and $n_{\tau}=\tau/\Delta$ is the target delay measured in frames.

Unlike the original Voxtral formulation, which places \texttt{[W]} according to the word offset, we anchor \texttt{[W]} to the word onset. The word's subword tokens are placed immediately after \texttt{[W]}, while all remaining positions are filled with the padding token \texttt{[P]}. This onset-based placement makes the ASR and turn state predictions available earlier relative to the spoken word. Because each word occupies one \texttt{[W]} position followed by at least one subword position, its representation requires at least two 80-ms steps.


Turn state targets are constructed using the same placement procedure. We first prompt a powerful language model~\cite{qwen} to assign a turn state label to each word. The resulting word-level label is then assigned to all target positions occupied by the corresponding \texttt{[W]} token and subword tokens. All unoccupied positions are labeled as \texttt{<|idle|>}, including positions corresponding to pre-speech silence, inter-word pauses, mid-utterance pauses, and trailing silence. In this way, the ASR and turn state targets are aligned on the same 80-ms discrete timeline.

This ASR-anchored supervision scheme has three main properties. First, each word-level turn state label is placed at the same positions as its corresponding ASR tokens. The turn head therefore predicts the state from the same decoder representations used by the ASR head to emit the associated transcription. Second, silence is explicitly supervised using the \texttt{<|idle|>} state, allowing a downstream interaction policy to perform endpointing by counting consecutive idle frames following a \texttt{<|complete|>} prediction. Third, because the ASR and turn state targets are aligned position by position rather than at the utterance or chunk level, both tasks are optimized over the same frame-synchronous discrete timeline.

\begin{table*}[t]
\centering
\caption{System-level results on bilingual Full-Duplex-Bench. $\uparrow$/$\downarrow$ indicate higher-/lower-better. Best per language in \textbf{bold}; second-best \underline{underlined}.}
\label{tab:fdb-overall}
\setlength{\tabcolsep}{3.6pt}
\renewcommand{\arraystretch}{1.12}
\small
\resizebox{0.82\textwidth}{!}{%
\begin{tabular}{@{}l c cc c cc ccc cc@{}}
\toprule
\multirow{2}{*}{Model}
  & Pause
  & \multicolumn{2}{c}{Turn-Taking}
  & Backchannel
  & \multicolumn{2}{c}{Interruption v1}
  & \multicolumn{3}{c}{Interruption v1.5}
  & \multicolumn{2}{c}{Overall} \\
\cmidrule(lr){2-2}\cmidrule(lr){3-4}\cmidrule(lr){5-5}\cmidrule(lr){6-7}\cmidrule(lr){8-10}\cmidrule(lr){11-12}
 & TOR$\downarrow$
   & TOR$\uparrow$ & RL$\downarrow$
   & RsR$\uparrow$
   & TOR$\uparrow$ & RL$\downarrow$
   & RpR$\uparrow$ & SL$\downarrow$ & RL$\downarrow$
   & ACC$\uparrow$ & Latency$\downarrow$ \\
\midrule
\multicolumn{12}{l}{\textit{English}} \\
dGSLM
    & 0.935 & 0.975 & 0.352 & -- & 0.917 & 2.531 & -- & -- & -- & 0.653 & 1.442 \\
PersonaPlex
    & 0.623 & \textbf{0.992} & \textbf{0.070} & -- & \textbf{1.000} & \underline{0.400} & -- & -- & -- & 0.790 & \textbf{0.235} \\
Moshi
    & 0.983 & 0.941 & \underline{0.265} & 0.060 & \textbf{1.000} & \textbf{0.257} & 0.500 & 1.160 & 1.470 & 0.504 & 0.788 \\
Freeze-Omni
    & 0.562 & 0.336 & 0.953 & 0.800 & 0.867 & 1.409 & 0.720 & 1.420 & 1.350 & 0.632 & 1.283 \\
Gemini Live
    & \underline{0.283} & 0.655 & 1.301 & \textbf{0.930} & 0.891 & 1.183 & 0.330 & 2.200 & 2.620 & 0.705 & 1.826 \\
GPT-4o
    & -- & -- & -- & 0.700 & -- & -- & \underline{0.780} & \textbf{0.230} & 1.500 & 0.740 & 0.865 \\
SoulX-Duplug
    & 0.352 & \underline{0.933} & 0.511 & 0.740 & \underline{0.970} & 0.773 & 0.770 & \underline{0.450} & \underline{1.030} & \underline{0.812} & \underline{0.691} \\
\textbf{X2-Turn (Ours)}
    & \textbf{0.224} & 0.807 & 0.901 & \underline{0.925} & 0.935 & 0.659 & \textbf{0.855} & 0.574 & \textbf{0.942} & \textbf{0.860} & 0.769 \\
\midrule
\multicolumn{12}{l}{\textit{Chinese}} \\
Freeze-Omni
    & 0.042 & 0.652 & \underline{0.780} & 0.700 & 0.975 & \textbf{0.232} & 0.440 & 1.300 & 1.630 & 0.745 & 0.986 \\
SoulX-Duplug
    & \underline{0.038} & \textbf{0.994} & \textbf{0.767} & \underline{0.800} & \underline{0.994} & \underline{1.089} & \underline{0.830} & \textbf{0.380} & \underline{1.150} & \underline{0.916} & \textbf{0.847} \\
\textbf{X2-Turn (Ours)}
    & \textbf{0.033} & \textbf{0.994} & 0.888 & \textbf{0.980} & \textbf{1.000} & 1.165 & \textbf{0.932} & \underline{0.578} & \textbf{0.979} & \textbf{0.975} & \underline{0.903} \\
\bottomrule
\end{tabular}%
}
\end{table*}

\section{Experimental setup}
\label{sec:exp_setup}
\subsection{Data Preparation}
\label{ssec:dataset_setup}

The corpora used in this work consist of two parts: Chinese-English ASR data and turn-taking data.  For the ASR data, we use AISHELL 1$\sim$4~\cite{aishell1,aishell2,aishell3,aishell4}, AliMeeting~\cite{alimeeting}, WenetSpeech~\cite{wenetspeech}, KeSpeech~\cite{kespeech}, LibriSpeech~\cite{librispeech}, GigaSpeech~\cite{gigaspeech}, TED-LIUM~\cite{tedlium}, and VoxPopuli~\cite{voxpopuli}, totaling approximately 26k hours (14k hours in Chinese and 12k hours in English). This portion of data is used in Stage 1 to strengthen the model's ASR capability in both languages.  
For the turn-taking data, we select a subset of the EasyTurn training set for Chinese (approximately 126 hours) and a subset of Fisher~\cite{fisher} telephone conversations for English (approximately 249 hours). This portion of data is used in Stage 2 for joint ASR and turn-taking modeling.  

For both parts of the data, Qwen3-ForceAligner~\cite{qwen3asr} was employed to obtain word-level timestamps. In addition, for the turn-taking data, we use Qwen3.5-Plus as an LLM annotator to perform word-level semantic turn state labeling, following the annotation criteria defined in Section 2.3.  Word-level labels are then projected onto the 80\,ms frame delayed-stream positions of their ASR word markers \texttt{[W]} and subword tokens.

\begin{table*}[t]
\centering
\caption{Module-level results on EasyTurn. $\mathrm{ACC}_{\mathrm{comp}}$, $\mathrm{ACC}_{\mathrm{incomp}}$,
and $\mathrm{ACC}_{\mathrm{bc}}$ denote utterance-level accuracy for the
complete, incomplete, and backchannel categories, respectively
($\mathrm{ACC}_{\mathrm{bc}}$ reported for Chinese only, as English has
no backchannel split). ``--'' denotes an unsupported or unavailable
state, and $\mathrm{latency}_{\mathrm{vad}}$ denotes the front-end VAD
delay incurred by cascaded methods.}
\label{tab:main_results}
\resizebox{0.85\textwidth}{!}{%
\begin{tabular}{l c cccc}
\toprule
\textbf{Method} & \textbf{Streaming} &
$\textbf{ACC}_{\mathrm{comp}}$ (\%, $\uparrow$) &
$\textbf{ACC}_{\mathrm{incomp}}$ (\%, $\uparrow$) &
$\textbf{ACC}_{\mathrm{bc}}$ (\%, $\uparrow$) &
\textbf{Latency} (ms, $\downarrow$) \\
\midrule
\multicolumn{6}{l}{\textit{Chinese}} \\
Paraformer + TEN Turn & {\color{red}\ding{55}} & 86.67 & 89.30 & -- & latency$_{\mathrm{vad}}$ + 204 \\
Smart Turn V3 & {\color{red}\ding{55}} & 91.33 & 60.00 & -- & latency$_{\mathrm{vad}}$ + 24 \\
EasyTurn & {\color{red}\ding{55}} & 96.33 & 97.67 & 91.00 & latency$_{\mathrm{vad}}$ + 263 \\
SoulX-Duplug & {\color{green!60!black}\ding{51}} & 89.33 & 79.33 & -- & 295 \\
\textbf{X2-Turn (Ours)} & {\color{green!60!black}\ding{51}} & \textbf{91.00} & \textbf{93.00} & \textbf{96.00} & 288 \\
\midrule
\multicolumn{6}{l}{\textit{English}} \\
SenseVoice En + TEN Turn & {\color{red}\ding{55}} & 95.60 & 76.59 & -- & latency$_{\mathrm{vad}}$ + 57 \\
Smart Turn V3 & {\color{red}\ding{55}} & 78.93 & 72.24 & -- & latency$_{\mathrm{vad}}$ + 21 \\
SoulX-Duplug & {\color{green!60!black}\ding{51}} & 77.67 & 88.96 & -- & 205 \\
\textbf{X2-Turn (Ours)} & {\color{green!60!black}\ding{51}} & \textbf{92.10} & \textbf{84.60} & -- & 225 \\
\bottomrule
\end{tabular}%
}
\end{table*}

\subsection{Implementation Details}
\label{ssec:implementation}

We use Voxtral-Mini-4B-Realtime~\footnote{\scriptsize{https://huggingface.co/mistralai/Voxtral-Mini-4B-Realtime-2602}} as the pretrained streaming backbone. All experiments fully fine-tune both the causal audio encoder and the language decoder. Training proceeds in two stages, with the ASR streaming delay $\tau$ sampled per batch between 1 and 30 frames (80--2400\,ms) in both stages, so that a single model covers a range of latency configurations.

During Stage 1, streaming ASR adaptation adapts the backbone to the delayed-stream ASR protocol on a large-scale Chinese--English corpus with frame-level ASR labels. Stage 2 then performs joint ASR and turn state fine-tuning. A turn state head is added, initialized as a copy of the ASR head, and the full model is fine-tuned on the Chinese--English turn-taking training set with paired ASR and frame-level turn state labels. The joint objective uses $\lambda = 0.1$.


\subsection{Latency Metric}
\label{ssec:latency-metric}
We measure \emph{latency} relative to the end of each word. For a word spanning 
$[s_i, e_i]$, the corresponding turn state becomes available at time $s_i + \tau$, i.e., after the configured streaming delay $\tau$ from the word onset. The resulting latency relative to the end of the word is 
\begin{equation}
L_i \;=\; \tau - (e_i - s_i),
\label{eq:latency}
\end{equation}
which measures how long after the user finishes speaking word $i$ the corresponding 
turn decision becomes available. If the word duration exceeds $\tau$, $L_i$ can be negative, meaning the state is available before the word ends. We report the average $L_i$ over all words in the EasyTurn test set. For cascaded baselines, following prior work~\cite{soulx}, we report inference time plus the front-end VAD delay $\mathrm{latency}_{\mathrm{vad}}$. While not strictly identical to $L_i$, this reflects their end-to-end decision delay.

\section{Results and analysis}
\label{sec:res}
We evaluate the proposed method at two levels. Section~\ref{ssec:fdb_result} reports system-level full-duplex behavior on bilingual Full-Duplex-Bench. Section~\ref{ssec:easyturn_result} then isolates the streaming turn module on EasyTurn. Sections~\ref{ssec:onset_offset} and~\ref{ssec:comparison_baseline} ablate word-boundary placement and the streaming delay $\tau$. Section~\ref{ssec:asr_comparison} reports streaming ASR quality.

\subsection{Main Results}
\label{ssec:main_result}

\subsubsection{System results on bilingual Full-Duplex-Bench}
\label{ssec:fdb_result}
We build a cascaded full-duplex dialogue system by coupling X2-Turn, which handles speech transcription and dialogue state management, with Qwen2.5-7B-Instruct~\cite{qwen2.5} for response generation and Qwen3-TTS-Streaming ~\footnote{\scriptsize{https://github.com/X-Square-Robot/Qwen3TTS-Streaming}} for speech synthesis. We evaluate the complete system on Bilingual Full-Duplex-Bench~\cite{fdb,fdbv1.5}, with results summarized in Table~\ref{tab:fdb-overall}. The gains are distributed across tasks rather than concentrated in one. Pause TOR is the lowest among all systems in both languages, indicating that X2-Turn takes the floor least often during intra-turn pauses. Backchannel RsR is highest in Chinese and second only to Gemini Live in English. Interruption RpR is highest in both languages, while interruption v1 TOR is competitive but not top-ranked. Chinese turn-taking TOR matches SoulX-Duplug.

The remaining gap lies in English turn-taking TOR and a slightly higher overall latency relative to SoulX-Duplug. SoulX-Duplug takes the floor as soon as the model emits \texttt{<|complete|>}, applying no additional decision rule, which yields very low response latency. We instead require a short run of \texttt{<|idle|>} frames following \texttt{<|complete|>} before taking the turn. This conservative endpointing rule reduces false take-overs during pauses, at the cost of lower English turn-taking TOR and a small added delay.

Latency remains competitive overall. X2-Turn avoids both the extreme pause TOR of Moshi and the multi-second interruption delays of Gemini Live. Among systems with fully reported metrics, X2-Turn therefore achieves the strongest overall ACC, with latency close to that of SoulX-Duplug.

\subsubsection{Module results on bilingual EasyTurn}
\label{ssec:easyturn_result}
Table~\ref{tab:main_results} compares X2-Turn (with $\tau=480\,ms$) against baselines on EasyTurn in terms of turn state classification accuracy and latency. We compare the last non-idle predicted state against the ground-truth utterance.

Among fully streaming systems, X2-Turn outperforms SoulX-Duplug on both accuracy metrics in Chinese, with similar latency (slightly lower). In English, X2-Turn achieves substantially higher complete accuracy (92.10 vs.\ 77.67) but lower incomplete accuracy (84.60 vs.\ 88.96), at a latency cost of 20ms. It also requires no auxiliary ASR model at inference: a single model jointly decodes transcripts and turn states.

Cascaded, VAD-dependent systems can post higher complete and incomplete accuracy, especially EasyTurn on Chinese, because they classify a full VAD segment. That segmentation adds a front-end delay that is not included in the raw inference times~\cite{flexduo}, so the higher accuracy does not imply a streaming responsiveness advantage. X2-Turn remains fully streaming and improves Chinese backchannel accuracy over EasyTurn (96.00 vs.\ 91.00).

\subsection{Onset vs.\ Offset Word-boundary Placement}
\label{ssec:onset_offset}
Voxtral places \texttt{[W]} at the word offset; we instead anchor it at
the onset (Section~2). Because the turn state is aligned with
\texttt{[W]}, this choice also sets when the decision can first appear.
A delayed-stream model waits \(\tau\) after that anchor. Offset
therefore cannot emit until the word has been fully heard plus
\(\tau\), whereas onset can decide while the speaker is still finishing
the word. The two cannot be compared at the same \(\tau\).
Table~\ref{tab:onset_offset} matches them on latency after the word
ends (Eq.~\ref{eq:latency}), using a larger \(\tau\) for onset.

\begin{table}[t]
\centering
\caption{Onset vs.\ offset \texttt{[W]} placement on the EasyTurn testset. Offset uses a smaller \(\tau\) to match latency after the word ends. Since we do not define an explicit \texttt{<|wait|>} label, \(\mathrm{ACC}_{\mathrm{wait}}\) is computed using semantic completeness together with the \texttt{<|complete|>} state.}
\label{tab:onset_offset}
\resizebox{1\columnwidth}{!}{%
\begin{tabular}{lccccccc}
\toprule
Placement & \(\tau\) (ms)
  & \(\mathrm{ACC}_{\mathrm{comp}}\)
  & \(\mathrm{ACC}_{\mathrm{incomp}}\)
  & \(\mathrm{ACC}_{\mathrm{bc}}\)
  & \(\mathrm{ACC}_{\mathrm{wait}}\)
  & Avg. & Lat.\ (ms) \\
\midrule
\multicolumn{8}{l}{\textit{Chinese}} \\
Offset & 240 & 90.33 & 91.67 & 93.00 & 96.00 & 91.88 & \textbf{240} \\
Offset & 320 & \textbf{93.33} & 88.67 & 92.00 & 96.00 & 91.75 & 320 \\
\textbf{Onset}  & 480 & 91.00 & \textbf{93.00} & \textbf{96.00} & \textbf{98.00} & \textbf{93.25} & 288 \\
\midrule
\multicolumn{8}{l}{\textit{English}} \\
Offset & 240 & 87.74 & 83.61 & --    & --    & 85.74 & 240 \\
Offset & 320 & 86.48 & 80.27 & --    & --    & 83.47 & 320 \\
\textbf{Onset}  & 480 & \textbf{92.10} & \textbf{84.60} & --    & --    & \textbf{88.49} & \textbf{225} \\
\bottomrule
\end{tabular}%
}
\end{table}

Onset is stronger overall in both languages and earlier in Chinese.
Offset, having heard the full word, slightly improves Chinese complete
accuracy; onset is better on incomplete, backchannel, and wait, i.e.,
less likely to take the floor too soon. On English, offset underperforms onset at both delays, and extra
delay after the word end does not close the gap. The modest drop from
240\,ms to 320\,ms may also reflect training variance. We therefore
keep onset in the X2-Turn system.

\begin{table}[h]
\centering
\caption{Ablation on the delay $\tau$ on the EasyTurn testsets.}
\label{tab:ablation}
\resizebox{0.95\columnwidth}{!}{%
\begin{tabular}{cc cccc}
\toprule
& $\tau$ (ms) & $\mathrm{ACC}_{\mathrm{comp}}$  & $\mathrm{ACC}_{\mathrm{incomp}}$  & Avg.  & Latency (ms) \\
\midrule
\multirow{4}{*}{Chinese}
& 480 & 91.00 & 93.00 & 92.00 & 288 \\
& 400 & 88.70 & 94.30 & 91.50 & 208 \\
& 320 & 87.33 & 94.00 & 90.67 & 128 \\
\midrule
\multirow{3}{*}{English}
& 480 & 92.10 & 84.60 & 88.49 & 225 \\
& 400 & 85.20 & 85.30 & 85.25 & 145 \\
& 320 & 82.70 & 87.60 & 85.09 & 65 \\
\bottomrule
\end{tabular}%
}
\end{table}

\begin{table}[t]
\centering
\caption{Streaming ASR word error rate (WER, \%, $\downarrow$).
Uni-ASR uses a 320\,ms chunk with beam search; Freeze-Omni uses a
chunk size of 4. ``--'' denotes unreported results.}
\label{tab:wer_comparison}
\setlength{\tabcolsep}{3.0pt}
\resizebox{\columnwidth}{!}{%
\begin{tabular}{@{}lccccc@{}}
\toprule
Test set
  & Uni-ASR~\cite{uniasr}
  & Freeze-Omni~\cite{freeze_omni}
  & \multicolumn{2}{c}{Stage1}
  & Stage2 \\
\cmidrule(lr){4-5}
 & & & 480\,ms & 2400\,ms & 480\,ms \\
\midrule
\multicolumn{6}{l}{\textit{Chinese}} \\
AISHELL-1                 & 2.90 & 2.79  & 2.57  & 1.48  & 3.94 \\
AISHELL-2 Android         & --   & --    & 4.70  & 3.42  & 5.70 \\
AISHELL-2 iOS             & --   & --    & 4.41  & 3.23  & 5.36 \\
AISHELL-2 Mic             & --   & --    & 4.54  & 3.54  & 5.55 \\
AISHELL-3                 & --   & --    & 3.16  & 2.13  & 4.66 \\
AISHELL-4                 & --   & --    & 18.72 & 16.42 & 22.79 \\
WenetSpeech Meeting  & --   & 14.2  & 9.25  & 7.68  & 12.18 \\
WenetSpeech Net      & --   & 12.6  & 9.50  & 8.40  & 10.39 \\
\midrule
\multicolumn{6}{l}{\textit{English}} \\
GigaSpeech                & --   & --    & 12.23 & 10.87 & 12.55 \\
LibriSpeech Clean    & 3.21 & 4.05  & 2.40  & 1.54  & 3.30 \\
LibriSpeech Other    & 7.71 & 10.48 & 5.87  & 3.77  & 8.53 \\
TED-LIUM                  & --   & --    & 4.62  & 3.66  & 4.50 \\
VoxPopuli                 & --   & --    & 9.16  & 6.05  & 11.91 \\

\bottomrule
\end{tabular}%
}
\end{table}

\subsection{Effect of Streaming Delay $\tau$ for Turn-taking}
\label{ssec:comparison_baseline}
We report the effect of different values of the streaming delay $\tau$ on both turn-taking and ASR performance. We first examine the effect of $\tau$ on turn-taking performance, considering three configurations (320, 400 and 480\,ms), as turn-taking inherently favors low-latency settings.

Table~\ref{tab:ablation} shows a consistent latency--accuracy trade-off across both languages. As $\tau$ decreases from 480\,ms to 320\,ms, latency drops substantially, while turn state accuracy degrades only mildly. This indicates that our model is robust under tight latency constraints, allowing a suitable operating point to be chosen for different latency requirements with little loss in turn-taking accuracy.

\subsection{Streaming ASR Performance Comparison}
\label{ssec:asr_comparison}

In this subsection, we compare our frame-synchronous approach against representative chunk-based streaming ASR systems. Since SoulX-Duplug does not report ASR results, we instead select two strong chunk-based streaming baselines, Uni-ASR and Freeze-Omni, for comparison. Table~\ref{tab:wer_comparison} summarizes the results. Under comparable or lower delay settings, Stage1-ASR at $\tau{=}480$\,ms already outperforms both baselines across nearly all evaluation sets. This indicates that frame-wise prediction with a short lookahead is competitive with, and often superior to fixed-chunk streaming even before accounting for its lower latency. Even after Stage2-Turn joint training, where turn-state supervision introduces a set-dependent degradation relative to Stage1-ASR at the same delay, our model still matches or exceeds Freeze-Omni. This suggests that the frame-synchronous backbone retains most of its recognition advantage over chunk-based streaming even under multi-task optimization.

\section{Conclusion}
This paper presents X2-Turn, a frame-synchronous dual-head extension of a pretrained delayed-stream ASR model for joint streaming ASR and turn state prediction. A parallel turn state head shares causal decoder representations with the ASR head, with ASR-anchored supervision projecting word-level turn labels onto the native 80\,ms token timeline. The streaming delay $\tau$ offers a controllable trade-off between turn-taking accuracy, response latency, and ASR quality. Experiments on bilingual EasyTurn and Full-Duplex-Bench demonstrate that X2-Turn achieves accurate turn-taking detection while maintaining low latency. Future work will further balance the ASR and turn state objectives and improve robustness in more challenging conversational settings.

\bibliographystyle{IEEEbib}
\bibliography{mybib}

\end{document}